\documentclass[preprint,12pt]{elsarticle}

\usepackage{algpseudocode}
\usepackage{amsmath}
\usepackage{amssymb}
\usepackage{amsthm}
\usepackage{boldtensors}
\usepackage{csquotes}
\usepackage{geometry}
\usepackage{graphicx}
\usepackage{lineno}
\usepackage{microtype}
\usepackage{multirow}
\usepackage{subcaption}
\usepackage{url} 
\usepackage{xcolor}

\usepackage[ruled,vlined]{algorithm2e}
\usepackage[font=footnotesize]{caption}

\usepackage{xcolor}

\journal{Journal of Computational and Applied Mathematics}

\begin{document}

\begin{frontmatter}



\title{Spatio-temporal Multivariate Cluster Evolution Analysis for Detecting and Tracking Climate Impacts}


\author[1]{Warren L. Davis IV}
\author[2]{Max Carlson}
\author[2]{Irina Tezaur\corref{cor1}}
\ead{ikalash@sandia.gov}
\cortext[cor1]{Corresponding author}
\author[1]{Diana Bull}
\author[1]{Kara Peterson}
\author[1]{Laura Swiler}

\affiliation[1]{organization={Sandia National Laboratories},
            city={Albuquerque},
            state={NM},
            country={USA}}
\affiliation[2]{organization={Sandia National Laboratories},
            city={Livermore},
            state={CA},
            country={USA}}

\begin{abstract}
Recent years have seen a growing concern about climate change and its impacts.
While Earth System Models (ESMs) can be invaluable tools for studying the impacts of 
climate change, the complex coupling 
processes encoded in ESMs and the large amounts of data produced by these models, together with the high internal variability of the Earth system, can obscure important source-to-impact relationships.  This paper presents a novel and efficient unsupervised data-driven approach for detecting statistically-significant impacts and tracing spatio-temporal source-impact pathways in the climate through a unique combination of ideas from anomaly detection, clustering and Natural Language Processing (NLP).  Using as an exemplar the 1991 eruption of Mount Pinatubo in the Philippines, we demonstrate that the proposed approach is capable of detecting known post-eruption impacts/events.   We additionally describe a methodology for extracting meaningful sequences of post-eruption impacts/events by using
NLP to efficiently mine frequent multivariate cluster evolutions, which can be used to confirm or discover the chain of physical processes between a climate source and its impact(s).  
\end{abstract}



\begin{keyword}
anomaly detection, clustering, multivariate data mining, natural language processing, Mount Pinatubo, climate impacts
\end{keyword}

\end{frontmatter}



\section{Introduction} \label{sec:intro}

Climate impacts with ramifications as serious as displacement, loss of livelihood and/or threat to health require an improved understanding of their causes.  In a complex system like the climate system, these impacts are normally produced through multiple interacting circumstances (natural variability, local features, and source forcings) and variables.  However, traditional detection and attribution (DA) methodologies, which identify a deviation outside of natural variability (detection of impact) and then determine the cause of that impact (attribution) have typically been designed to only associate one source with one impact \cite{hegerl2010}.  Methods designed to transition through a series of steps by which a complex evolution can be traced could greatly enhance our ability to understand high consequence ramifications with legal, political, and national security implications \cite{Burger2020}. 

\subsection{Contributions} \label{sec:intro_contribs}

This paper presents a novel data-driven approach for detecting and tracking spatio-temporal impacts within the climate system, and is based on two ingredients: (i) unsupervised 
signature-based clustering, and (ii) multivariate data-mining. 
 The former approach, signature-based clustering, relies on compressed representations of the data 
 known as \textit{signatures} \cite{SheadNovel2023, Aditya:2019, Ling:2017}, which are
 defined in  
 spatially-local analysis partitions and grouped together using $k$-means clustering \cite{K-means}.
 By performing statistical tests that determine whether the resulting clusters 
 are statistically different from analogous clusters defined for a baseline dataset, 
 we are able to detect signals/impacts that are above the internal variability of the underlying data.  
  The latter approach, multivariate data-mining, enables not only impact detection, but also the tracing of the chain of physical processes that link climate sources and their impacts -- referred to herein as \textit{pathways} or \textit{source-impact pathways}.  In the data-mining step of our workflow, we construct, for each analysis partition, a time-history of 
  the cluster to which the analysis partition belongs, and utilize Natural Language Processing (NLP) \cite{NLP} to 
efficiently mine the data for frequent and physically-meaningful multivariate cluster evolutions. From these sequences, we derive spatio-temporal ``pathways'' between a climate source and its impact, providing climate scientists a powerful tool for confirmatory or exploratory analysis of climate data.

We develop and demonstrate our approach on a volcanic eruption exemplar, the 1991 eruption of Mount Pinatubo in the Philippines.  This eruption 
produced aerosols that encircled the globe and modified the scattering of incoming shortwave radiation from the sun as well as the absorption of outgoing radiation from the Earth, leading to a surface cooling and stratospheric warming event that lasted up to two years following the eruption \cite{McCormick:1992}.  We select this exemplar for two primary reasons: (i) the eruption was large enough in magnitude to produce primary impacts that rose above the internal variability of the global climate system \cite{hansen1992potential, kilian2020impact, parker1996impact, self1993atmospheric}, and (ii) it possesses a well-characterized spatio-temporal evolution governed by stratospheric circulation patterns. 
While our approach can be applied to simulated and/or observed data, herein we utilize simulated data generated by running a modified version of the U.S. Department of Energy's Energy Exascale Earth System Model (E3SM) \cite{Golaz:2022} that includes prognostic stratospheric aerosols and is known as E3SMv2-SPA \cite{Brown:2024}.  We demonstrate that our methods are able to correctly identify and track impacts related to the stratospheric warming 
effect known to have been caused by the Mount Pinatubo eruption, and that these impacts are statistically different from natural climate variability present in a baseline ``Without Pinatubo'' counterfactual simulation in which the eruption does not occur.

\subsection{Related previous work} \label{sec:intro_litreview}

Discovering connections between a climate source 
and its impact(s) 
is an ongoing research challenge in climate science \cite{IPCCar6ch3}.  
Most state-of-the-art evaluation techniques operate by positing a set of source-impact pairs
and then employing standard statistical methods on simulated and/or observed data to infer connective relationships between the posited source-impact pair \cite{hegerl2010,IPCCar6ch3}. 
Often, such methods, such as Pearson’s correlation coefficient \cite{Pearson:2024}, inappropriately assume a linear relationship.  Due to the innate complexity of the climate, these methods typically require large amounts of data to overcome the sources of variability and uncover complex, often nonlinear, correlations.  Additionally, most popular methods are limited to single variable data, which can miss important variable interactions present in a source-impact trajectory.

Fingerprinting is the most common DA technique in climate and it looks to establish spatial and/or temporal patterns of impacts under forced conditions \cite{Hasselmann1993,santer1993}.  Although this has been subject to much development over the years \cite{AllenStott2003,ribes2013,Wills2020} and has even pushed towards regional studies \cite{Bonfils2008,stott2010}, it is still fundamentally designed to work with a single source-impact pair.  Multivariate fingerprinting has been employed \cite{Bonfils2020,Marvel:2020}, but these studies are not exploiting the series of steps by which a complex evolution can be traced.

Recently there has been a focus on developing multivariate models, sometimes sensitive to spatial and temporal characteristics, that enable a representation of the relationships between variables. Causal inference has been proposed as a mathematical framework for addressing why an impact has occurred, but has only recently focused on climate \cite{Nowack:2020, Pearl:2009, Spirtes:2016}. Typically, causal inference approaches start with a full representation between variables and then attempt to remove causal paths that are not significant.  
It has been documented by Runge \textit{et al.} \cite{Runge:2019} that there are many issues with current causal methods which must be addressed for Earth systems applications, including high-dimensionality, scalability of causal methods, unobserved variables, contemporaneous effects and feedback cycles, and long time-scales. 
Other approaches based on operator neural networks \cite{Hart:2023,Hart:2024}, echo-state networks augmented with feature importance \cite{Goode:2024,Ries:2024,McClernon:2024}, spatially-resolved causal methods (CaStLe) \cite{Nichol:2024}, and multivariate space-time dynamic models \cite{Garrett:2024} have shown promise capturing and predicting the evolution of downstream variables from a series of input variables.  Methods like conditional multi-step attribution \cite{Wentland:2024} and inverse optimization \cite{Hart:2024} have even established frameworks by which the source magnitudes can be determined from the impact by employing multiple variables in the pathway from source to impact. 
However, developing descriptions of source-impact trajectories and employing them in attribution is still a relatively novel pursuit. 

While the ingredients underlying our approach -- clustering and NLP-based data-mining -- are well-established, 
our unique combination of these two tools is fundamentally new and unlike the existing state-of-the-art methods summarized above.  
As our approach is inherently unsupervised, it does not require large amounts of training data.  Additionally, while most popular methods are limited to single variable data, which can miss important variable interactions present in a source-impact trajectory, our approach can efficiently handle multiple variables or features of interest. 
Finally, while most existing approaches are \textit{confirmatory}, that is, they can only confirm posited relationships, our method has the potential to actually \textit{discover} unknown relationships present in a dataset. Doing so, however, is currently computationally expensive. Therefore, the rest of this paper will focus on purely on the model creation, pattern mining, and confirmatory steps. 

Various researchers have studied the application of clustering algorithms to climate data. Specifically, research in \cite{Pampuch2023} applied Ward's method, $k$-means, and self-organizing maps to temperature and precipitation features separately, towards determining homogeneous regions across South America in terms of these variables. Our research performs clustering on each feature separately as well, but applies an additional methodology to combine the features, producing multivariate clusters that can be used for downstream pathway analysis. In \cite{Franzke2011}, researchers utilize clusters and hidden Markov models to find different regimes of behavior within a particular climatological phenomenon. Although they study different simulations, the purpose of the study is not to define differences in climate impact due to specific events or forcings, but rather to make sure the regimes they identify are stable across initial conditions. We employ a different process for determining cluster stability that focuses on the ability to discretize features consistently throughout a simulation and across different initial conditions, and does not insist that transitions between clusters within a simulation occur infrequently for the clusters to exist separately. Below, we describe this methodology in detail, and specifically develop techniques to identify salient climate impacts that result from specific forcings and initial conditions.

As for employing clustering analyses to study climate impact, we are aware of a several recent papers. One utilizes $k$-means cluster analysis to determine the relative roles 
of forcing and initial conditions in a set of ensembles simulating the climate response of the Tambora eruption in 1815 \cite{Zanchettin:2019}. 
The research in \cite{Zanchettin:2019} uses the difference between a simulation's individual surface temperature anomalies and the ensemble‐mean anomaly as a descriptor for the simulation, then uses $k$-means clustering to cluster the simulations into three clusters. Their analysis then describes the conditions in which simulations within a particular forcing regime (Best, High, and Low) are either cohesive or scattered amongst different clusters. In contrast, our approach first clusters the individual state variables and assigns each state variable for each partition and each timestep to a particular cluster for that feature, then applies statistical tests to find statistically-significant anomalies on these clustering assignments. 
Moreover, unlike our approach, the method in \cite{Zanchettin:2019} is not designed to perform either signal detection or spatio-temporal pathway tracking. 
Another recent study uses hierarchical clustering to find a geographical region similarity based on ozone and temperature burden \cite{Jahn2022}. While these methods do, in fact, cluster based on climate impact, the focus is not on differentiating entire simulations due to climate events or forcings, but rather to differentiate geographical regions based on the state of the climate variables-- grouping regions with similar ozone and temperature burdens. In contrast, our research seeks to find areas that are being affected by a particular \emph{climate dynamic}, regardless of the similarity in climate variables. We seek areas that are being similarly affected by a particular event or forcing, regardless of whether they share similar feature/state variables.



In addition, researchers performing clustering on data that change over time typically seek to understand how the nature of individual clusters is changing in distribution (e.g., dataset shift), and use methods that can accommodate the development of new clusters and removal of ones that are less used, such as in \cite{cig2016step}. That capability is counter to what we want to achieve in this instance, as we 
are looking for clusters that have consistent meaning, and additionally want to describe the state of a simulation in terms of the clusters (and cluster transitions) that are present. Closer analogs to our research include the works  \cite{cataldi2010emerging} and \cite{berrocal2013reina}, which seek to identify connections between terms used in text corpora.  These methods differ from our approach in two ways: (i) they are working with words -- data that are already in a discrete space -- whereas we are interested in climate variables, which are usually in a continuous space, and (ii) they seek to find communities of related terms (words that are all referring to one particular ``topic''), whereas our goal is to identify multivariate transitions over time that we can use as descriptors to differentiate simulations that vary due to climactic events.

\subsection{Organization} \label{sec:intro_org}
The remainder of this paper is organized as follows.  Section \ref{sec:pinatubo} describes our Mount Pinatubo exemplar problem, the E3SMv2 code used to simulate this scenario, and the data utilized in our analyses.  In Section \ref{sec:clustering}, we detail our signature-based clustering methodology and demonstrate its utility for climate impact detection in the context of our exemplar problem, focusing primarily on single variable analyses.  Next, in Section \ref{sec:mining}, we introduce our multivariate data-mining approach, which provides a mechanism for handling multivariate relationships in a computationally tractable way, and enables the identification of pathways by tracking cluster evolution patterns in time and by mining these data for physically-meaningful cluster evolution sequences.  A concluding summary and discussion of future work is given in Section \ref{sec:conc}. 




\section{Climate Exemplar: the 1991 Eruption of Mount Pinatubo in the Philippines Simulated Using E3SM} \label{sec:pinatubo}

\subsection{Mount Pinatubo Eruption} \label{sec:pinatubo_desc}

The methods presented herein are demonstrated on a climate exemplar problem: the June 15, 1991 eruption of Mount Pinatubo in the Philippines.  This eruption was the largest eruption of the satellite era and the second largest eruption in the twentieth century \cite{Newhall:1997}, injecting 5-10 Tg of sulfur \cite{Rieger:2020}, primarily as sulfur dioxide (SO$_2$) gas, into the stratosphere, where the gas reacted with water to form a hazy layer of sulfate aerosol particles 
\cite{Robock:2020}.   The aerosols spread zonally 
according to the stratospheric winds dictated by the phase of the Quasi-Biennial Oscillation (QBO, which was westerly near the tropopause \cite{Thomas2009,Ehrmann:2024})
within the 20$^{\circ}$S to 30$^{\circ}$N latitude band in the lower stratosphere (20--27km) in the first few weeks after the eruption \cite{McCormick:1992,bluth1992}.  
The aerosol optical depth is a combined measure of mass loading and effective radius of all particles in the atmosphere.  Tropical aerosol optical depth peaked at 0.35 in October and November 1991 \cite{stenchikov1998} but transport poleward with the stratospheric Brewer-Dobson Circulation was much slower.  After approximately 6 months the aerosols reached the poles \cite{ramachandran2000,stenchikov1998} resulting global average aerosol optical depth depths of around 0.15 in early 1992  \cite{russell1996,stenchikov1998}.  Stratospheric aerosol levels returned to pre-Pinatubo levels in 1993 \cite{ramachandran2000}. 

As a consequence of the increase in aerosol optical depth, incoming shortwave radiation was scattered reducing shortwave forcing by 4 W/m$^2$ at the top of atmosphere \cite{allan2014, liu2015} resulting in a cooling of the troposphere by $\sim$0.4K between June 1992 and October 1992~\cite{Kremser:2016, ramachandran2000}.  Additionally, these stratospheric aerosols absorbed outgoing longwave and near-infrared radiation to resulting in a global reduction of 2.5--3 W/m$^2$ in August of 1992 \cite{allan2014, liu2015}. The absorption of outgoing longwave terrestrial radiation in the stratosphere warmed the lower levels ($\sim$14--22km) globally by 2-3K near the end of 1992~\cite{ramachandran2000}. 
The event is additionally believed to have caused other global climatic impacts, 
including decreases in precipitation \cite{gillett2004, TrenberthDai2007, Gu2007}, and global sea-level \cite{church2005}, as well as increase in diffusivity of incoming radiation \cite{Robock:2020, proctor2018} 

The Mount Pinatubo eruption is an ideal starting point for developing and verifying methods aimed at detecting and tracing climatic impacts, due to the large signal-to-noise ratio in some of the relevant climate variables such as aerosol optical depth and stratospheric temperature, as the well characterized spatio-temporal evolution of processes following the eruption.

\subsection{Simulations Using the E3SM} \label{sec:pinatubo_e3sm}

The data utilized in our numerical experiments were generated by running simulations using an enhanced fork\footnote{Available at: \url{https://github.com/sandialabs/CLDERA-E3SM}.} of version 2 of the U.S. Department of Energy's Energy Earth System Model (E3SMv2) \cite{Golaz:2022}, which we describe succinctly here.  
Typically, ESMs such as
E3SMv2 prescribe the 
location and magnitude 
of absorption and scattering of stratospheric aerosol from explosive volcanic
eruptions (i.e., take them from an input file), rather than simulating them explicitly within the
 model. In order to 
 enable a dynamically consistent reproduction of 
 the impacts from Mount Pinatubo's eruption
 and enable sensitivity studies with respect to various characteristics of this eruption, it was necessary for us to implement a prognostic volcanic aerosol capability in the code to evolve SO$_2$ gas from the eruption into sulfate aerosols.  This new implementation, known as E3SMv2-Stratospheric Prognostic Aerosols (E3SMv2-SPA), modifies aerosol microphysics
to accurately simulate stratospheric volcanic aerosols and was validated against observational data
\cite{Brown:2024}. Longer-term coupled climate simulations performed using E3SMv2-SPA confirmed that the
changes to the stratospheric aerosols did not change the average climate in E3SMv2.

Once E3SMv2-SPA was implemented and verified, we performed an ensemble of limited variability (LV) simulations of the Mount Pinatubo eruption using this model 
as described in more detail in \cite{Ehrmann:2024}. Limited variability was achieved by matching historical conditions for the QBO and the El Ni\~no Southern Oscillation (ENSO) as characterized by changes in tropical Pacific sea surface temperatures. These two major modes of natural variability respectively precondition the direction of travel of the injected SO$_2$ gas in the lower stratosphere as well as temperature and precipitation values globally \cite{Davey2014}.  
The LV ensemble was generated by perturbing the 
temperature by values near machine precision, 
which led to slightly different 
synoptic 
behavior 
at the time of the eruption due the chaotic nature of the climate system. These simulations were run from June 1, 1991 to December 31, 1998, and as described in \cite{Brown:2024}
introduced a point injection of 10 Tg SO$_2$ at 5.15$^{\circ}$N and 120.35$^{\circ}$E, at an altitude of 18-20km, which
lasted 6 hours. The eruption took place on June 15, 1991, so as to mimic as closely as possible
the actual Mount Pinatubo eruption. More details about these simulations can be found in \cite{Brown:2024, CLDERA_SAND, Ehrmann:2024, ehrmannSAND2024}.

\subsection{Data} \label{sec:pinatubo_data}

The present analysis was based on an ensemble having five LV members, which we will refer to herein as the ``With Pinatubo'' simulations.  For comparison purposes, we also had at our disposal a
paired five member set of ``Without Pinatubo'' counterfactual ensembles, containing no volcanic eruption but using
the same initial conditions as the ensembles containing the eruption.  
All simulations were performed using a 1$^{\circ}$
configuration of E3SMv2-SPA, which has a resolution of 110 km with 72 vertical layers for the atmosphere, and a resolution of 165 km for land.  

For the sake of brevity and to enable method verification, 
we focus our attention herein on a well-understood impact of the Mount Pinatubo eruption, namely the so-called ``stratospheric warming pathway'', in which the formation of sulfates increases aerosol optical depth, which, in turn,
increases longwave radiation absorption, and leads to an increase in the stratospheric temperature \cite{labitzke1992,Brown:2024}.  The three variables relevant to this pathway and used in our analysis are denoted by ``AEROD\_v'', ``FLNT'' and ``T050'', and summarized in Table \ref{tab:e3sm_vars}.  These represent the aerosol optical depth, the net longwave radiation and the stratospheric temperature.  All three variables are two-dimensional fields; the T050 variable is evaluated at 50 HPa, and whereas the AEROD\_v and FLNT variables are column-integrated.  When running E3SMv2-SPA, it is possible to output these and other variables at different temporal frequencies.  The results presented herein all utilized average daily data for the three variables of interest, and worked with the data remapped onto a latitude-longitude (lat-lon) grid with a small region near the poles trimmed off where data is missing at various times of the year.  We select to focus our analysis on the stratospheric warming pathway rather than the surface cooling pathway described earlier as the stratospheric temperature variable, T050, is 
subject to much less variability in the climate system 
than the
surface temperature variable.  Repeating our analysis on the surface cooling pathway would be an interesting future research endeavor.

\begin{table}[hbt!]
    \centering
     \caption{E3SMv2-SPA stratospheric warming pathway variables.}
    \begin{tabular}{cccc}
    \hline \hline
     \textbf{Variable}& \multirow{2}{*}{\textbf{Long Name}} &\multirow{2}{*}{\textbf{Units}} & \multirow{2}{*}{\textbf{Description}}\\
     \textbf{Name}&  &&\\
     \hline
         \multirow{3}{*}{AEROD\_v} & Aerosol  & \multirow{3}{*}{$-$} & 
         Column-integrated aerosol \\
         & optical & & optical depth (missing \\
         & depth&&data in polar winter)\\\hline
         \multirow{3}{*}{FLNT} &  Top of atmosphere  & \multirow{ 3}{*}{W m$^{-2}$} & Longer wavelength outgoing\\
          & net longwave && radiation preferentially absorbed  \\ &radiation flux&       & by sulfate aerosols in stratosphere  
         \\
          \hline
         T050 & Temperature at 50 hPa & K & 
         Stratospheric temperature\\\hline
        \hline
    \end{tabular}
    \label{tab:e3sm_vars}
\end{table}







\section{Impact detection via signature-based clustering} \label{sec:clustering}

\subsection{Methodology} \label{sec:clustering_methods}

We begin by describing our approach for detecting impacts in spatio-temporal 
climate data using \textit{signature-based clustering}.  This methodology consists of the following key ingredients: (i) splitting the spatial grid into 
\textit{analysis partitions} (groups of cells or spatial regions), 
(ii) calculating \textit{signatures} (reduced representations 
of the data) in each analysis partition, (iii) performing a \textit{$k$-means clustering} 
on the signatures, and (iv) applying statistical tests (e.g., $t$-tests) to detect statistically-significant features/impacts within the clustered data.  



\subsubsection{Analysis partitions and signatures}

\begin{figure}[!htb]
  
  \centering
  \includegraphics[scale=0.5]{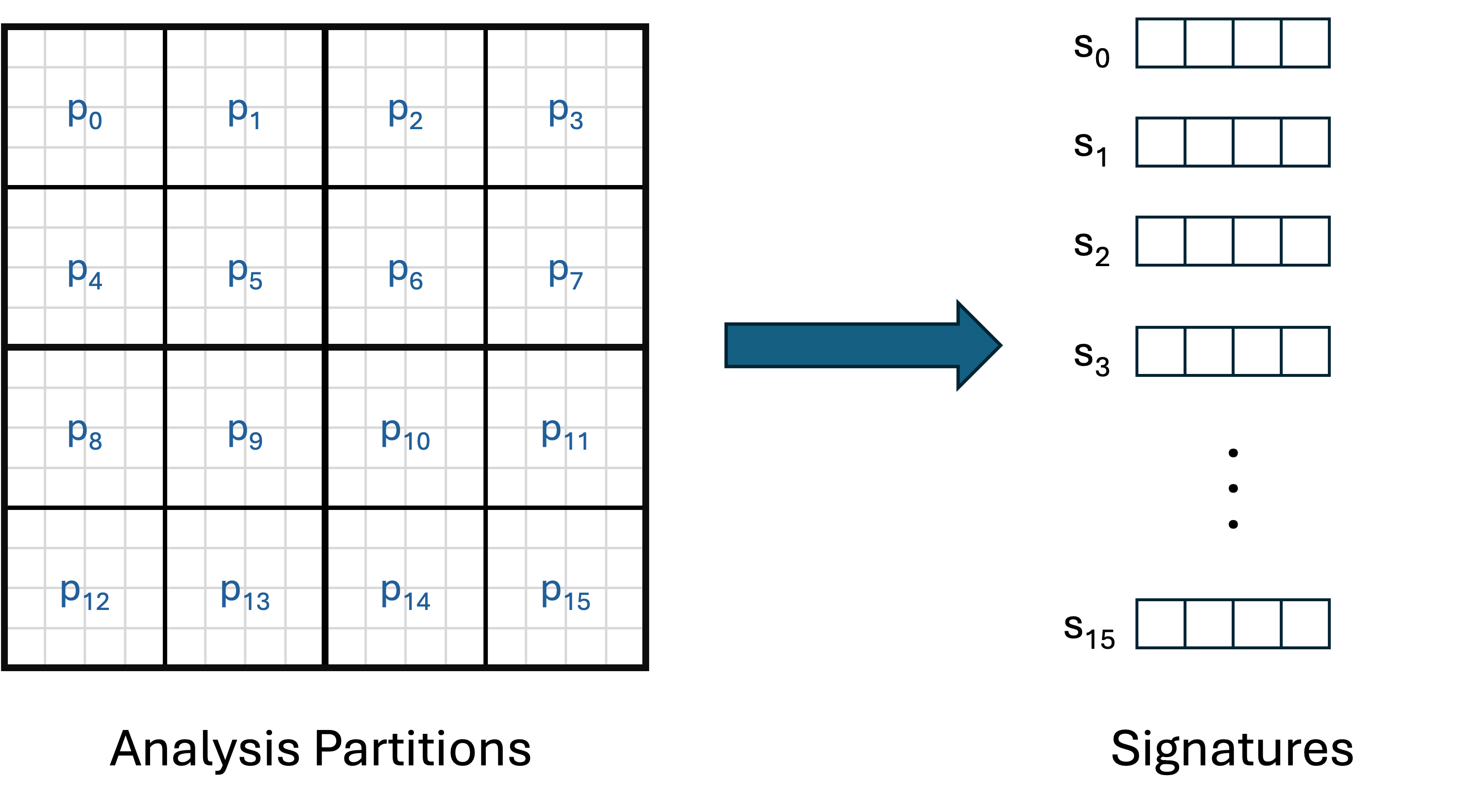}
  \caption{Variables in each analysis partition $p_i$ (left panel) are converted to a reduced rank/dimension signature $s_i$ (right panel).} \label{fig:signatures}
\end{figure}

The concept of analysis partitions and signatures was originally 
conceived within a communication-minimizing unsupervised framework for detecting anomalies (or ``interesting events") in HPC simulations, developed under a project known as {\tt ISML} (In-Situ Machine Learning for anomaly detection) \cite{SheadNovel2023, Aditya:2019, Ling:2017}.  The core {\tt ISML} algorithm, which serves as a starting point for the present work,  starts by dividing the spatial grid underlying an HPC simulation into \emph{analysis partitions}: groups of cells or mesh points that fit completely onto a processor or node (left panel of Figure \ref{fig:signatures}).  As will become clear shortly, selecting partitions with partition sizes larger than one cell improves 
the computational efficiency of the {\tt ISML} algorithm, and effectively smooths the underlying simulation data.

Once analysis partitions are defined, the information in each analysis partition is then compressed using one of many techniques delineated in \cite{SheadNovel2023, Aditya:2019, Ling:2017}. These compressed representations, called \emph{signatures} (right panel of Figure \ref{fig:signatures}), contain less data than the original partition, making them 
cheaper and easier to communicate across processors without requiring massive bandwidth.  The simplest example of a signature is a variable average over an analysis partition.  In this case the data inside a partition are reduced down to a single number.  Other commonly-used signatures are quartile and percentile, although we note more sophisticated signatures based on feature importance values \cite{Ling:2017}, feature moment metrics \cite{Aditya:2019}, the singular value decomposition (SVD), principal component analysis (PCA), etc. are also possible.

\subsubsection{Clustering} \label{sec:clustering_metrics}

In {\tt ISML}, signatures were used to detect anomalies.  This was accomplished by comparing signatures across partitions using pre-specified \textit{measures}, and evaluating these measures using \textit{decision} functions to determine which partitions are likely to contain outliers.  In contrast, one of the main goals of the present research is to identify partitions that contain similar climate profiles within a given dataset/simulation. To enable this, once partitions are selected and signatures are computed within each partition, we apply the $k$-means clustering algorithm \cite{K-means} to the signature data over the course of the entire simulation for all simulations. We use the entire timeline of the simulations to ensure that we have a consistent frame of reference, ensuring that definitions of what is ``average'' do not shift over time. Likewise, we use both the With Pinatubo and Without Pinatubo simulations to ensure that the cluster semantics are consistent, regardless of any exogenous events.

The $k$-means algorithm is a natural fit for clustering. It is based on the Euclidean distance, which works fine for single variable values. We have no \emph{a priori} knowledge about the minimum core points or neighborhood distance, so an algorithm like {\tt DBScan} would require an additional investigation to determine those values. We opt instead to perform an analysis that optimizes cluster stability. The end result is a model that generates cluster assignments for each analysis partition.

 We determine the ``optimal" number of clusters $k$ by utilizing a common cluster stability metric called the \textit{Adjusted Rand Index (ARI)}. Although the $k$-means algorithm is deterministic, the outcome of a clustering produced by this algorithm will depend on the initial starting points for the cluster centers, known as \emph{centroids}.
 If the assignments of observations to clusters vary wildly depending on the initial starting point, this suggests a lack of distinguishable structure within the feature space. 
 We argue here that selecting algorithmic parameters such that a stable 
 set of clusters is produced is of primary importance, both for rigor and interpretability.
In addition to this, we seek to have enough clusters to adequately distinguish between different clusters of climate variables, but not so many as to be confusing during interpretation. 
To achieve this balance, we use as a metric the ARI, which helps us determine stability with respect to the number of clusters $k$ and other algorithmic parameters.

The Rand Index is an metric used to evaluate the similarity between two clusterings of data \cite{rand1971objective}. A Rand Index of 1 indicates that the two clusterings are identical, whereas a Rand Index of 0 indicates that the two clusterings have nothing in common. 
The ARI is similar, except that it takes into account that similarities between clusterings can happen purely by chance \cite{hubert1985comparing}. The ARI is bounded between $-0.5$ and $1$ \cite{chacon2023minimum}, with negative values indicating that two clusterings are more different than would be expected by random chance.  An ARI score of 1 is optimal, as it implies that the same clustering is obtained every time clustering is performed (up to a permutation of cluster IDs).  


In order to estimate the cluster stability for a given value of $k$ (or any other user-specified parameter), we perform the following experiment. We first generate a clustering using `$k$-means++' initialization \cite{Arthur:2007} and iterate ten times over various centroid seeds. The $k$-means clustering algorithm seeks to minimize intra-cluster distance, the distance between elements in each cluster. Therefore, we choose the centroid initialization that results in the minimum intracluster distance as the `near-optimal' standard of comparison. 
We then generate six auxiliary clusterings with random initial centroids. For each of these clusterings, we compute the ARI between each randomly initialized clustering and the near-optimal clustering. We average all of the individual ARIs, and this \emph{averaged near-optimal ARI} becomes the final cluster stability estimate. This experiment is repeated for whatever range of $k$ (or any other user-specified parameter) that we are interested in for a given variable. 

\subsubsection{Cluster analysis and statistical tests}

Once clustering has been performed, the resulting clusters can be used to extract 
additional information 
relevant to signal detection.

While the cluster IDs produced by the $k$-means algorithm are typically assigned in 
an arbitrary way, each cluster's center can be queried for the average signature value 
within that cluster.  This information can be used to impose an interpretable 
order on the cluster IDs.  Additionally, in an offline analysis setting like the one 
considered herein, 
we can also compute statistics on a cluster-by-cluster basis using the simulation data to further characterize the members of a cluster having a given cluster ID.   

For some parts of our analysis, it can be useful to compute the number of partitions that belong to 
each cluster at each timestep of the simulation. These time series give us a high level view of the behavior of each cluster over time.  As shown in Section \ref{sec:clustering_results}, it is possible to use this information to detect an impact: in the case of the Mount Pinatubo eruption, we see a sharp increase in the number of clusters having a higher stratospheric temperature and a simultaneous decrease in the number of clusters having a lower stratospheric temperature once the aerosols from the eruption have had a chance to spread across the globe, thereby warming the stratosphere.  
In addition to tracking the total number of partitions within a cluster, it can be useful to also look at the percentage of the globe that is covered by cells belonging to a particular cluster at a given time.  This information can provide insight into which regions are experiencing a given impact at a given time. 

In order to identify spatio-temporal signals/impacts in our clustered data, we rely on statistical $t$-tests.  
These tests require a baseline simulation or ensemble of simulations.  
In the present work, we use as a baseline an ensemble of Without Pinatubo simulations (simulations in which 
there is no Mount Pinatubo eruption; see Section \ref{sec:pinatubo}), 
so that each With Pinatubo simulation has associated with it a Without Pinatubo ensemble member.  For each pair of simulations, the data are clustered and time series of the number of partitions that belong to each cluster ID are created.  We then perform a $t$-test to determine if the means of these two sets of time series are likely to come from
different distributions.  If the $t$-test reveals that they do, this provides evidence that the differences between
the two datasets are due to the eruption, rather than internal climate variability.


\subsubsection{Single variable vs. multivariate analysis} \label{sec:clustering_single_vs_multi}



In the simplest application of our method, all the variables of interest are handled independently, that is, they are clustered independently and statistical tests are applied to each variable separately.  The downside of this approach is that it does not take into account variable interactions.  
As our climate simulations include multiple variables, we need to develop a way to use our clustering techniques in this context. 

Multivariate clustering can be complicated if the individual variables differ significantly. For instance, some variables may be on the scale of 0 to 1, while other variables may have values in the thousands. If using a clustering technique based on Euclidean distance, such as $k$-means,  the larger variable will have a greater impact on the clustering results. 

To mitigate this problem, variables are often scaled or standardized so that all variables lie within the same range. However, this standardization does not account for the semantics of a variable (one variable may be more important than others for an application), nor do they account for the fact that the variables may be of different distribution (e.g., one variable may have a Gaussian distribution, while another may be uniform or bimodal).
Many methods in the literature approach this problem by creating a unified representation of the variables and then developing a similarity metric over the unified representation that takes in to account the nature of the different variables \cite{abbdullin2012clustering, chu2024hybrid}. The basic idea is to create analytically, or derive through techniques such as \textit{similarity learning}, a proper distance measure.

Herein, we take a different approach that preserves the distinct characteristics of each variable clustering and does not require learning a proper similarity measure.  First, we cluster each variable independently, determining the ``optimal'' number of clusters using measures of the cluster stability (Section \ref{sec:clustering_metrics}). 
Next, each partition will then receive a cluster assignment for each individual variable and for each independent time-step. We then simply treat the cluster assignments for each partition as a tuple of the cluster assignments for each variable. For example, if a partition at a particular time-step belongs to cluster 3 for the first variable, cluster 1 for the second variable, and cluster 4 for the third variable, its cluster tuple at that time-step would be listed as $(3,1,4)$. Each tuple is then tracked/analyzed as its own cluster.  A clear downside of this approach is that 
it quickly leads to an exponentiation in the number of cluster tuples that need to be analyzed.  We handle this difficulty using the data mining approach described later, in Section \ref{sec:mining}.  

\subsection{Results/analysis} \label{sec:clustering_results}

We now present results from applying our signature-based clustering approach 
to the exemplar problem of interest, the 1991 eruption of Mount Pinatubo in the Philippines, 
described earlier in Section \ref{sec:pinatubo}.  
 We focus our attention on the stratospheric warming temperature pathway, in which changes to aerosol optical depth (AEROD\_v) 
lead to an increase in longwave radiation absorption (FLNT), which leads to an increase in the stratospheric temperature (T050).  Hence, there are three variables of interest, AEROD\_v, FLNT and T050. Each of these clusters are clustered 
individually; the multivariate clustering case is discussed later in Section 
\ref{sec:mining}. 


\subsubsection{Single variable clustering} \label{sec:clustering_results_single_var}

We begin by determining the ``optimal'' values  of the three relevant parameters, the partition size, the signature type and the number of clusters $k$, for each of variable of interest.  Table \ref{tab:optimal_param_vals_single} summarizes the optimal values (based on the averaged near-optimal ARI; Section \ref{sec:clustering_metrics}) of our three parameters for each variable of interest. Note that in this table, multiple optimal values of $k$ meet our criteria but these values are those that were chosen specifically for the results in this paper.

\begin{table}[hbt!]
    \centering
\caption{Optimal parameter values for single variable clusterings}
\begin{tabular}{c||ccc}
 & AEROD\_v & FLNT & T050  \\
    \hline \hline
Partition size &  \multicolumn{3}{c}{$3 \times 3$} \\
\hline 
Signature & \multicolumn{3}{c}{percentile(5)} \\
\hline 
Number of clusters $k$ & 4 & 4 & 5\\
\end{tabular}
\label{tab:optimal_param_vals_single}
\end{table}

\begin{figure}[!htb]
  \centering
  \includegraphics[scale=0.7]{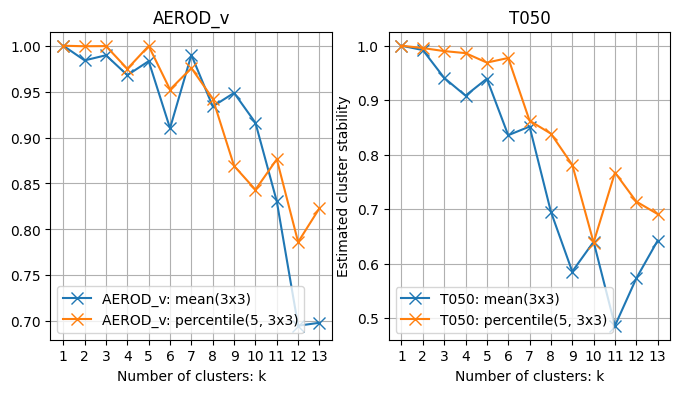}
  \caption{Comparison of cluster stability estimates for T050 and AEROD\_v when using mean or percentile(5) signatures. For both variables, percentile(5) signatures result in better stability in the range of $k$ we are interested in.  The cluster stability estimate is the averaged near-optimal ARI, discussed in Section \ref{sec:clustering_methods}.}
  \label{fig:mean_vs_percentile_stability}
\end{figure}

In determining these optimal values, we considered partition sizes ranging from $1 \times 1$ to $7\times 7$ grid cells, and cluster sizes $k$ between 1 and 12.  We also considered two candidate signatures: the mean and the percentile signature with 5 percentiles with ranges [0, 0.25, 0.5, 0.75, 1.0], referred to herein as percentile(5).
We first compared the estimated stability using mean and percentile(5) signatures with the partition size fixed to $3\times3$ as seen in Figure \ref{fig:mean_vs_percentile_stability}. The percentile(5) signature produced more stable clusterings compared to a simple mean signature, in particular for the stratospheric temperature (T050) variables.

\begin{figure}[!htb]
  \centering
  \includegraphics[scale=0.65]{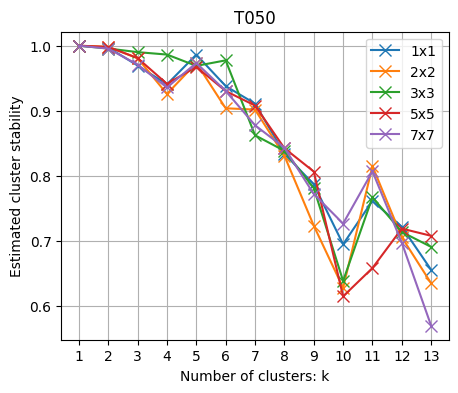}
  \caption{Comparison of how different partition sizes impact cluster stability for T050. Since partition size does not seem to have much of an impact on stability, it can be chosen to maximize data reduction while minimizing loss of spatial precision, which for this work is a $3\times3$ analysis partition.  The cluster stability estimate is the averaged near-optimal ARI, discussed in Section \ref{sec:clustering_methods}.
  }
  \label{fig:partition_size_stability}
\end{figure}

In order to determine the impact of partition size on cluster stability, we repeated this experiment by fixing the signature to percentile(5) and estimating the stability for partitions of size 
$1\times1$, $2\times2$, $3\times3$, $5\times5$, and $7\times7$. The results of this experiment can be seen in Figure \ref{fig:partition_size_stability}. The reader can observe
that the partition size does not have a large impact on stability, 
meaning we are free to choose partition size to maximize data reduction while simultaneously minimizing spatial precision loss. In the present work, we have chosen to use $3\times3$ analysis partitions.

\begin{figure}[!htb]
  \centering
  \includegraphics[scale=0.55]{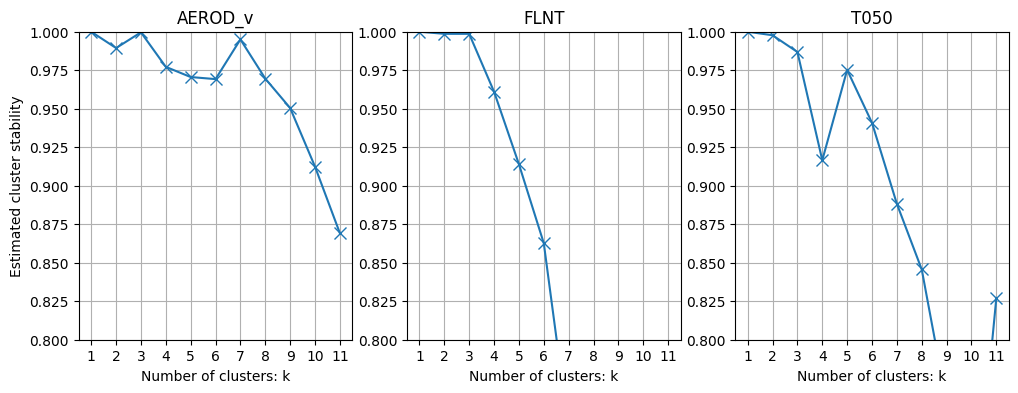}
  \caption{Cluster stability estimates for stratospheric heating pathway variables T050, FLNT, and AEROD\_v.  Candidate optimal $k$ values are chosen as the peaks in stability when $k>3$. Multiple choices of $k$ meet our selection criteria and the results can be seen in Table \ref{tab:optimal_param_vals_single}.  The cluster stability estimate is the averaged near-optimal ARI, discussed in Section \ref{sec:clustering_methods}.}
  \label{fig:all_vars_stability}
\end{figure}

With analysis partition size and signature chosen, we computed stability estimates for each of the three stratospheric heating pathway variables and the results can be seen in Figure \ref{fig:all_vars_stability}. We selected optimal $k$ values by choosing the first peak in stability after $k>3$. There is a secondary peak in AEROD\_v at $k=7$ which meets our selection criteria but for this paper, we have limited our choice to $k=4$. The optimal values of $k$ that we use for this work can be seen in Table \ref{tab:optimal_param_vals_single}.

Once optimal values of the parameters are chosen, 
the limited variability E3SM ensemble data are clustered simultaneously (for each case, With Pinatubo and Without Pinatubo) using the settings determined by the previous analysis. We do this for each variable in the stratospheric heating pathway to get the single variable clusterings T050[5], AEROD\_v[4], and FLNT[4] where the notation VAR[$k$] means a clustering of variable VAR with $k$ clusters. In order to avoid cumbersome cluster ID notation, we have come up with some descriptive names for each cluster ID that can be seen in Table \ref{tab:all_cluster_info} along with some statistics about what value ranges typically fall under each cluster ID.

\begin{table}[hbt!]
    \centering
\caption{Statistics and descriptive names for stratospheric heating pathway variable clusterings.}
\begin{tabular}{c||ccc}
AEROD\_v & Cluster ID & Descriptive Name & Value Range \\
\hline \hline
& 0 & Clear sky & $1.08\times 10^{-1}\pm3.75\times 10^{-2}$  W m$^{-2}$\\
& 1 & Low opacity & $2.33\times 10^{-1}\pm5.48\times 10^{-2}$ W m$^{-2}$\\
& 2 & Medium opacity & $4.70\times 10^{-1}\pm1.20\times 10^{-1}$ W m$^{-2}$\\
& 3 & High opacity & $1.04\pm3.80\times 10^{-1}$ W m$^{-2}$
\end{tabular}
\quad
\begin{tabular}{c||ccc}
FLNT & Cluster ID & Descriptive Name & Value Range \\
\hline \hline
& 0 & Least radiative flux & $181.31\pm17.22$ \\
& 1 & Less radiative flux & $215.33\pm14.42$ \\
& 2 & More radiative flux & $252.31\pm14.66$ \\
& 3 & Most radiative flux & $288.22\pm12.43$
\end{tabular}
\quad
\begin{tabular}{c||ccc}
T050 & Cluster ID & Descriptive Name & Value Range \\
\hline \hline
& 0 & Coolest & $202.15\pm3.48$K \\
& 1 & Cooler & $209.58\pm1.39$K \\
& 2 & Neutral & $213.79\pm1.39$K \\
& 3 & Warmer & $219.00\pm1.71$K \\
& 4 & Warmest & $225.49\pm2.69$K
\end{tabular}
\label{tab:all_cluster_info}
\end{table}

\begin{figure}[!htb]
  
  \centering
  \includegraphics[scale=0.5]{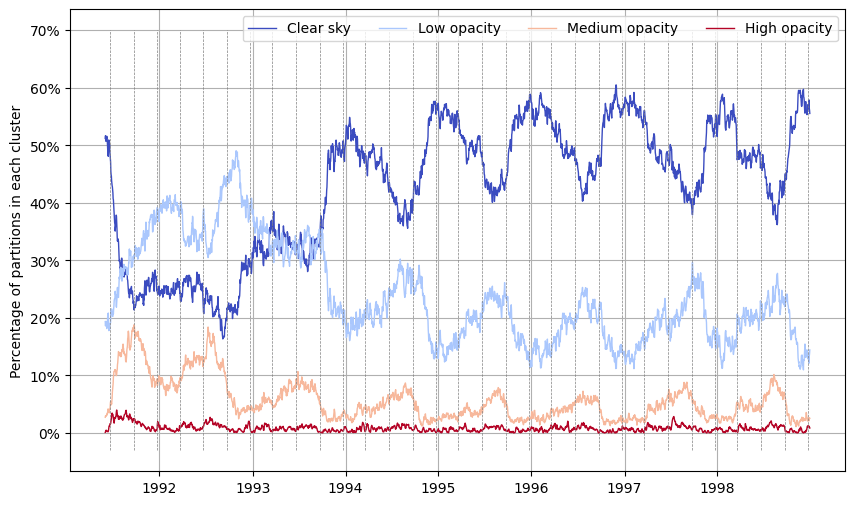}
  \caption{Examples of instantaneous cluster counts for a clustering of the AEROD\_v variable in a single E3SM simulation with total of four clusters.}
  \label{fig:global_cluster_counts}
\end{figure}

We begin our analysis of the single variable clusterings by showing a plot (Figure \ref{fig:global_cluster_counts}) of instantaneous cluster counts as a function of time for a clustering of AEROD\_v using the optimal parameter values from Table \ref{tab:optimal_param_vals_single}.  The reader can observe that, before the eruption, the clearest cluster (shown in dark blue) was the most prevalent.  However, after the eruption in the second half of 1991, there is a noticeable decrease in the number of partitions that are the most clear (dark blue) and a noticeable increase in the number of partitions that are have medium/high opacity (pink and red).  This shift occurs as a result of aerosol production and spread following the Mount Pinatubo eruption.
The time series of cluster membership over time can thus be used to detect changes in cluster membership due to the eruption, as will be discussed further in Section \ref{sec:mining}.

\begin{figure}[!htb]
  \centering
  \includegraphics[scale=0.45]{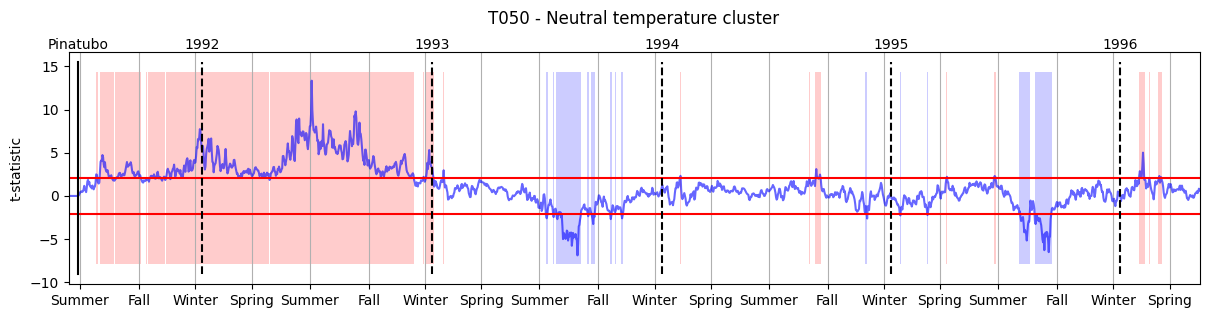}
   \caption{Time-history of the $t$-statistic for the T050 variable's neutral temperature cluster. When the $t$-statistic falls outside of the range denoted by the two horizontal red lines, this indicates a statistically-significant difference between the memberships of the With Pinatubo clustering and the Without Pinatubo clustering. A statistically-significant increase in With Pinatubo membership is noted by a red highlighting and a decrease with blue.}
   \label{fig:tstat_example}
\end{figure}

Next, we demonstrate how our approach can enable detection through a statistical comparison of the clustered data with respect to a baseline Without Pinatubo simulation.  The input data to this approach consists of two sets of paired five member ensembles: one with the Mount Pinatubo eruption and one without.  For each pair of simulations, we apply our clustering algorithm (Section \ref{sec:clustering_methods}) to the data, and create a time series of the number of partitions that belong to each cluster ID, as shown in Figure \ref{fig:global_cluster_counts}.  Next, treating the number of partitions in each cluster for the two (With Pinatubo and Without Mount Pinatubo)
five-member ensembles as samples from the same 
distribution, we conduct a $t$-test to determine if the means of the two sets of samples are statistically likely to come from different distributions.  
While the E3SMv2 runs are of high spatio-temporal resolution, the data at each grid cell is limited to our five ensemble members.  This limits the types of statistical analysis techniques that are applicable.
The result of this test gives us a $t$-statistic, plotted as a function of time in Figure \ref{fig:tstat_example} for the T050 variable, specifically for cluster 2 (the neutral temperature cluster) from which one can infer the  probability of how likely it was that the two sets of ensemble means came from different distributions. 
Moreover, we can use the sign of the difference to determine if a signal corresponds to statistically-significant increase (positive sign) or decrease (negative sign) in cluster's membership.  From Figure \ref{fig:tstat_example}, it can be inferred that the neutral temperature cluster has a nearly sustained increase in membership for as long as two years following the Mount Pinatubo eruption.

\begin{figure}[!htb]
  \centering
  \includegraphics[scale=0.55] {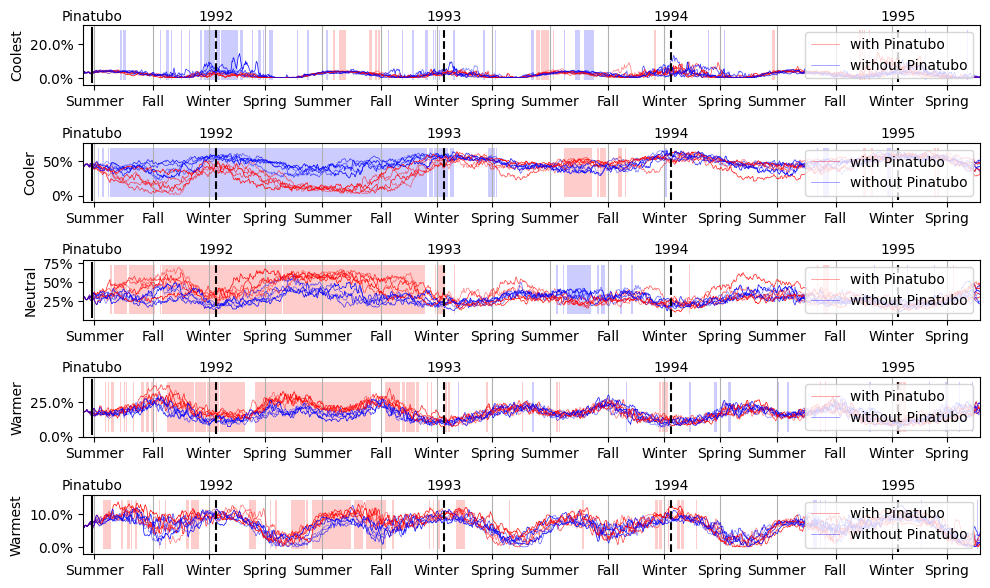}
   \caption{Statistical significance for all T050 clusters similar to Figure \ref{fig:tstat_example}. Instead of plotting the $t$-statistic, here the blue and red curves correspond to the percentage of analysis partitions that belong to each cluster ID. Red curves are for With Pinatubo ensemble members and blue curves are for Without Pinatubo ensemble members. Red and blue highlighting correspond to a statistically-significant increase or decrease in membership respectively.}
  \label{fig:tstat_5cluster_t050}
\end{figure}

In Figure \ref{fig:tstat_5cluster_t050}, we plot the cluster membership curves for each cluster in the limited variability ensemble for the stratospheric temperature variable, T050. In this figure, instead of plotting the $t$-statistic, we plot the cluster membership curves for the With Pinatubo and Without Pinatubo cases for all ensemble members. Time steps for which the cluster membership difference is detected to be statistically-significant are highlighted in either red or blue, corresponding to an increase or decrease, respectively.

From Figure \ref{fig:tstat_5cluster_t050}, we can see that the cooler temperature cluster has significantly reduced cluster membership sustained across the first two years after the Mount Pinatubo eruption. Correspondingly, the neutral, warmer and warmest clusters see significant increases during this same time period. This result is consistent with the idea of the kind of stratospheric heating we expect to see due to the Mount Pinatubo eruption and is captured by our single variable clustering and $t$-test detection technique.

\begin{figure}[!htb]
 
  \centering
  \includegraphics[scale=0.65]{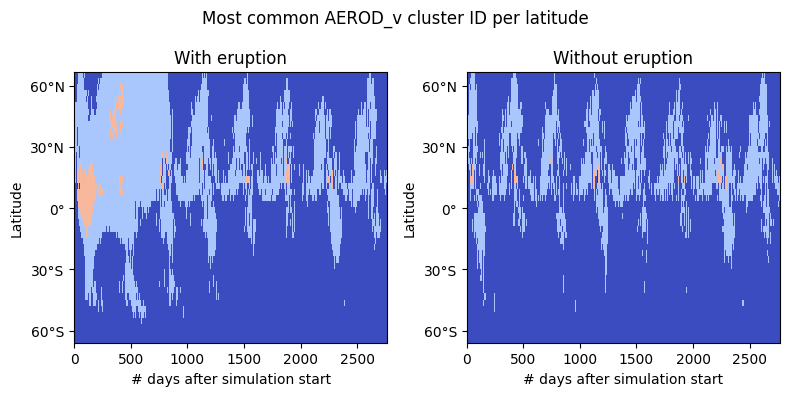}
   \caption{Most common cluster IDs per latitude for aerosol optical depth with and Without Pinatubo (E3SMv2-SPA limited variability ensemble). There are four total cluster IDs ranging from dark blue (clear sky) to dark red (high opacity aerosol optical depth); see Table \ref{tab:all_cluster_info} and Figure \ref{fig:global_cluster_counts}.}
  \label{fig:clusterology_aerod_v}
\end{figure}

In order to understand where and when we expect to see impacts due to elevated AEROD\_v, we used our single-variable clustering of AEROD\_v to produce a plot of the most common cluster ID found in each latitude over the course of the simulation.  
We plot this quantity in the left and right panels of Figure \ref{fig:clusterology_aerod_v} for the With Pinatubo and Without Pinatubo simulations, respectively.
In this figure, it can be seen that the AEROD\_v starts out highly concentrated around the eruption and stays mostly centered around the equator for the first 180 days and then spreads out up into the northern hemisphere. This pattern is noticeably absent from the Without Pinatubo data (right panel of Figure \ref{fig:clusterology_aerod_v}).  

It is interesting to note that there is a slight increase in aerosol optical depth that shows up in the southern hemisphere shortly after the Mount Pinatubo eruption in both the With Pinatubo and Without Pinatubo subplots.  This feature can most likely be attributed to another volcanic eruption that occurred shortly after the Mount Pinatubo eruption, namely the eruption of the Cerro Hudson volcano in Chile on August 8, 1991 \cite{Case:2017}.  The Cerro Hudson eruption was not only much smaller than the Mount Pinatubo eruption, but also further away from the equator; as a result, its impacts were more localized and shorter lived, as Figure \ref{fig:clusterology_aerod_v} shows.

\section{Pathway identification via multivariate data mining} \label{sec:mining}

It was demonstrated in Section \ref{sec:clustering} that, by performing a signature-based clustering on climate data and applying statistical tests to the data, it is possible to detect statistically-significant climate impacts.  
Here, we go one step further by developing a data mining-based approach that extracts pathways, defined as the interactions of a set of variables in space-time due to an external forcing, from time-series of clustered data.   This approach has the additional advantage of enabling analyses involving multivariate clusters 
in a tractable way.  


\subsection{Methodology}  \label{sec:mining_methods}

\begin{figure}[!htb]
  \centering
  \includegraphics[scale=0.45]{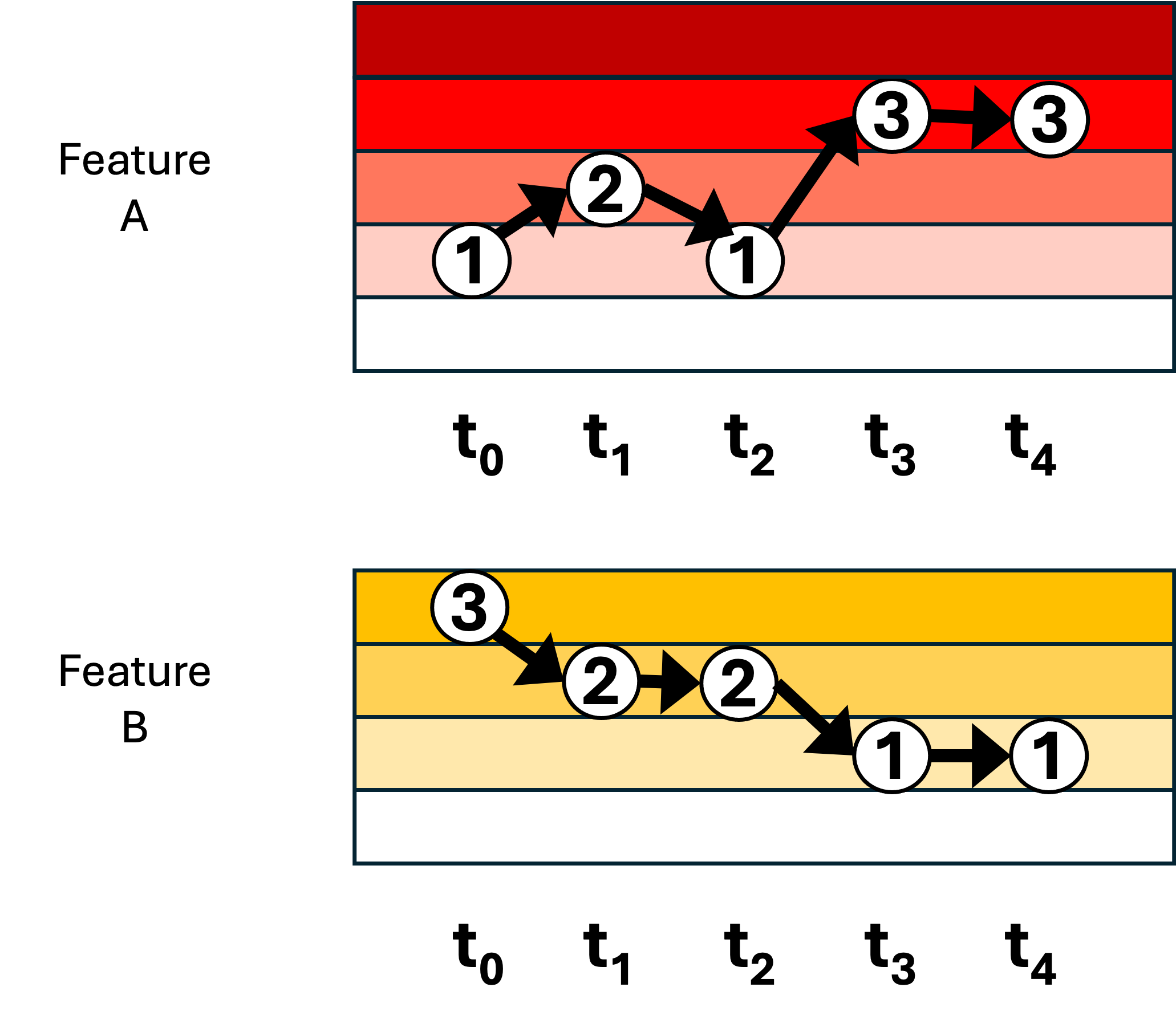}
   \caption{Notional example of two features of a partition evolving through cluster assignments over time.  The numbers in the white circles represent cluster IDs.}
  \label{fig:featureEvolution}
\end{figure}

Here, we take an Eulerian perspective to tracking cluster assignment changes. 
For any given analysis partition, the cluster assignment can change over time due to shifts in the values of the individual variables within the partition, as shown 
in Figure \ref{fig:featureEvolution} 
for a notional set of cluster change states for two variables (or features) in one particular partition.
In this example, 
Feature A begins in cluster 1 at timestep $t_0$, then changes to cluster 2 at $t_1$. It then proceeds to cluster 1 at $t_2$, then cluster 3 at $t_3$, where it remains through $t_4$. This can be written as an evolution of $1\rightarrow 2\rightarrow1\rightarrow3\rightarrow3$. The state of Feature B proceeds through its own evolution, likewise written as $3\rightarrow2\rightarrow2\rightarrow1\rightarrow1$. We can represent the evolution of two or more features over time by joining the individual cluster states into tuples. In this case, the joint evolution of Features A and B can be written as $(1,3)\rightarrow (2,2)\rightarrow(1,2)\rightarrow(3,1)\rightarrow(3,1)$.  
With representations such as these, we can now describe the evolution of every partition in a simulation over one or multiple variables.

The basic idea of our approach is to create evolution sequences for each set of variables in each cluster, and to search these sequences for subsequences of a specified length so as to find commonalities or differences in the prevalence of those subsequences.  Our hypothesis is that partitions undergoing similar climate dynamics will have sequences that share a similar prevalence. If a partition differs significantly in sequence prevalence from another partition, we would expect that that partition is experiencing different climate dynamics. Such differences might be observable in partitions taken from near the poles versus partitions taken from the tropics, or in sequences that happen before versus after some climate event.

The most expensive step of the workflow described above is the search step, as
it must be done on a partition-by-partition basis, and 
there can be exponentially many subsequences present in the cluster evolution sequence data.  We handle the subsequence search problem  efficiently using 
tools from NLP \cite{NLP}.

In NLP, there is the concept of an \emph{$n$-gram}, which is a specific sequence, of specified length \(n\), of symbols. Symbols can be defined many ways, such as letters, numbers, or even words. NLP techniques can perform varying analyses with these $n$-grams, such as assessing the statistical prevalence of a certain sequence of words in a corpus of documents. The number of unique $n$-grams increases potentially exponentially as $n$ increases, while the count of each unique instance generally decreases monotonically. As a result, $n$ is usually constrained to handle the more common cases. We utilize the Natural Language Toolkit ({\tt NLTK}) \cite{bird2009natural}, as the libraries available in this toolkit are specifically tailored to perform $n$-gram extraction efficiently.

In our research, each tuple of cluster assignments for every time step is treated as a symbol. For every partition, we record the symbols present at every time step. We then de-duplicate the symbols, such that successive symbols that are identical are compressed into just one instance. We then record the $n$-grams present throughout the entirety of the simulation, varying \(n\) from 2 to 4, to get an understanding of the underlying prevalence of types of climatological evolution. We also record the location and time for each of these $n$-grams.



Once $n$-grams of cluster evolution patterns have been recorded, 
we can aggregate the statistics of sequence prevalence over all partitions of a simulation and perform 
statistical analyses on these data to extract further information. 
Simulations with similar climate dynamics should be statistically-similar to each other, and vice versa.  For the Mount Pinatubo exemplar considered herein, it is natural to compare our With Pinatubo simulations to the Without Pinatubo analogs.


To improve interpretability, it is helpful to translate 
cluster evolution sequences into verbal descriptions. For instance, let us assume that Feature A in Figure \ref{fig:featureEvolution} is average temperature regimes, so that the cluster IDs represent the following temperature regimes: 0=coolest, 1=cooler, 2=neutral, 3=warmer, and 4=warmest, as in Table \ref{tab:all_cluster_info}. In this case, the aforementioned sequence can be described as follows:
\begin{displayquote}
The temperature of this partition starts off in a \emph{cooler} regime at $t_0$. The very next timestep, the partition heats up and becomes \emph{neutral}, but settles back to \emph{cooler} at $t_2$. At $t_3$, however, the partition suddenly becomes \emph{warmer}, and remains that way through the next timestep.
\end{displayquote}



While our NLP-based approach for mining cluster evolution patterns in our clustered data is capable of \textit{discovering} the relative prevalence of different cluster ID sequences, it is often helpful to use this approach in a \textit{confirmatory} setting, i.e., to postulate that a sequence corresponding to an expected physical process should be present in the data, and then to use our approach to confirm/deny this hypothesis.  This confirmatory approach greatly reduces the computational complexity of our algorithm and can help ensure that physically meaningful evolutions are extracted/analyzed. We focus on confirmatory analysis for the rest of this study.

\subsection{Results/analysis} \label{sec:mining_results}

Having described our approach, we now apply it to the Mount Pinatubo exemplar of interest herein (Section \ref{sec:pinatubo}).  

To verify our hypothesis about cluster evolution sequence differentiability, 
we worked with climate scientists to identify sequences of patterns 
that are expected to be more prevalent in our With Pinatubo ensembles
than in our Without Pinatubo ensembles.  
It is well accepted in the climate community that the Mount Pinatubo eruption resulted in a decreased longwave flux (FLNT), which in turn 
 led to an increase in the stratospheric temperature (T050) in partitions containing the injected aerosols (AEROD\_v). 
In order to apply our approach, we first translate this assertion into 
three simpler assertions, listed below. In this discussion, we assume that a 
partition having a smaller cluster ID corresponds to a lower value of the underlying variable or feature.
\begin{itemize}
    \item \textit{Assertion 1 (A1).} AEROD\_v must be non-zero for every step in the evolution.
    \item \textit{Assertion 2 (A2).} For FLNT, the end of the evolution must be in a cluster having a lower cluster ID than at the beginning of the evolution, and the FLNT cluster cannot increase during the evolution (monotonically non-increasing).
    \item \textit{Assertion 3 (A3).} For T050, the end of the evolution must be in a cluster having a higher cluster ID than the beginning of the evolution, and the T050 cluster cannot decrease during the evolution (monotonically non-decreasing).
\end{itemize}
We will use our approach to identify clusters that satisfy all three of these assertions, A1--A3.  

We begin by applying our multivariate clustering approach (Section \ref{sec:clustering_single_vs_multi}) to the AEROD\_v, FLNT and T050 variables using the optimal parameter values 
deduced earlier (Section \ref{sec:clustering_results_single_var}, Table \ref{tab:optimal_param_vals_single}).  
In our experiments, the first variable in the tuple is the cluster for AEROD\_v, the second is the cluster for FLNT, and the third element of the tuple is the cluster for T050. Therefore, the format we use to express the multivariate cluster is (AEROD\_v, FLNT, T050).

Having performed the clustering, we extract evolutionary patterns for these variables 
using the aforementioned multivariate representations. We allow for single patterns (cluster states where there is no evolution) as well as evolutionary patterns up to length 4 (3 total transitions from the initial state). 
These transitions can happen over any time frame contained within the simulation. We mine only patterns that are acyclic. If, for instance, cluster states 1 through 3 are acyclic and a potential cluster state 4 would cause a repeat of a previous state, we would record cluster states 1 through 3 as an evolutionary pattern of length 3, but these clusters appended with the fourth cluster would not be included in a new evolutionary pattern of length 4.

Across the three variables of interest, many unique combinations of cluster tuples and evolutions can be found. For instance, one of the more prevalent evolutions in Simulation 1 is:
\begin{equation} (0,0,3)\rightarrow(0,1,3)\rightarrow(0,1,2)\rightarrow(0,0,2)
\end{equation}

Looking at Table \ref{tab:all_cluster_info}, we can see that this is an evolution where the FLNT cluster starts off in the lowest cluster (Least radiative flux) and rises slightly to the second closer (Less radiative flux), then goes back down to the lowest cluster at the end, accompanied with the temperature starting at the second highest cluster (Warmer) and slightly reducing to the middle cluster (Neutral) over the cluster evolution, all while the aerosols are in the lowest cluster (Clear sky). This evolution involved 3396 partition time steps between both the with and Without Pinatubo scenarios. This is in contrast, to the evolution:
\begin{equation}
(0,1,1)\rightarrow(0,0,2)\rightarrow(0,1,2)\rightarrow(0,0,4).
\end{equation}
In this evolution, the FLNT cluster oscillates between the second lowest and lowest clusters, while the temperature rises (sharply so at the end), in the presence of clear sky. This somewhat erratic evolution occurred in only \textit{one} partition time step over both simulation instances. Gathering the statistics of these evolutions allows one to exercise choices, such as ignoring erratic climate events like this one or focusing on finding rare events, depending on the nature of the research application. Mining all the possible combinations of pattern evolutions is computationally expensive, and increases exponentially with the length of the pathway. The majority of the time for analysis is spent on this step, with the second highest expenditure being the downselect to fit the assertions (the criteria have to be checked for each pattern in every pathway) so we limit our evolutionary pathways to length 4 in the interest of feasibility.

 
In our research, we will be using evolution patterns to find a specific climate process in which we are interested. First, we downselect the evolutionary patterns to those that fit the three criteria listed above. 
Next, we look for these patterns in each ensemble member for both the Without Pinatubo and With Pinatubo simulations.
The results of this study are summarized in Table \ref{tab:EvolutionPartitionTime}, 
which gives the number of unique evolutions and partition-timestep pairs identified in each of our simulations, totaled across both the With Pinatubo and Without Pinatubo instance for that simulation index. The simulations had 7841 patterns found on average. 
In accordance with the climate science-informed assertions, the With Pinatubo simulations have on average 66\% more partition-timesteps with the sought-after evolution patterns than the Without Pinatubo simulations.  


\begin{table}[!htbp]
\caption{Evolution and partition-timestep counts for each Without Pinatubo and With Pinatubo simulation (i.e., ensemble member). Statistics for the averages are rounded to the nearest integer.}
\label{tab:EvolutionPartitionTime}
\centering
\begin{tabular}{cccc} \hline
\multirow{2}{*}{Simulation \#}  & \multirow{2}{*}{Unique Evolutions} & Without Pinatubo &   With Pinatubo
\\ & & Partition-Timesteps &  Partition-Timesteps      \\ \hline
1       & 7651  &   602,677 &   952,806 \\   
2       & 7670  &   572,589 &   977,741 \\
3       & 7998  &   580,999 &   978,765 \\
4       & 7979  &   560,845 &   100,3341 \\
5       & 7905  &   602,789 &   942,295 \\  \hline
Average & 7841  &   583,980 &   970,990 \\  \hline
\end{tabular}
\normalsize
\end{table}

Table \ref{tab:top_evolutions} lists the top five multivariate cluster evolution patterns of length 4 that fit assertions A1--A3 found in the simulations, as ranked by total partition-timesteps. They all feature a non-zero AEROD\_v cluster, a decreasing FLNT cluster, and an increasing T050 cluster. As can be seen, these cluster evolution pathways happen significantly more in the With Pinatubo simulations than the ones without, sometimes twice as much.

\begin{table}[!htbp]
    \centering
     \caption{Total partition-timestep counts for some of the top evolution pathways of length 4 for simulations Without Pinatubo and With Pinatubo.
     }
    \begin{tabular}{cccc}
    \hline
         \multirow{2}{*}{Pathway \#} & \multirow{2}{*}{Cluster Evolution Pathways} & Total Without & Total With  \\
        &  & Pinatubo & Pinatubo\\ \hline
         1 & $(1,3,1)\rightarrow(1,3,2)\rightarrow(1,2,2)\rightarrow(1,1,2)$&  15,659 & 31,600\\
         2 & $(1,2,1)\rightarrow(1,1,1)\rightarrow(1,0,1)\rightarrow(1,0,2)$&  12,129 & 14,946\\
         3 & $(2,3,1)\rightarrow(1,3,1)\rightarrow(1,3,2)\rightarrow(1,2,2)$&  10,397& 14,940\\
         4 & $(1,3,1)\rightarrow(1,3,2)\rightarrow(2,3,2)\rightarrow(2,2,2)$&  8,765 & 15,087\\
         5 & $(1,3,1)\rightarrow(1,3,2)\rightarrow(1,2,2)\rightarrow(1,0,2)$&  6,135& 13,979\\
    \end{tabular}   
    \label{tab:top_evolutions}
\end{table}

\begin{figure}
\centering
\begin{subfigure}{.8\textwidth}
  \centering
  \includegraphics[width=\linewidth]{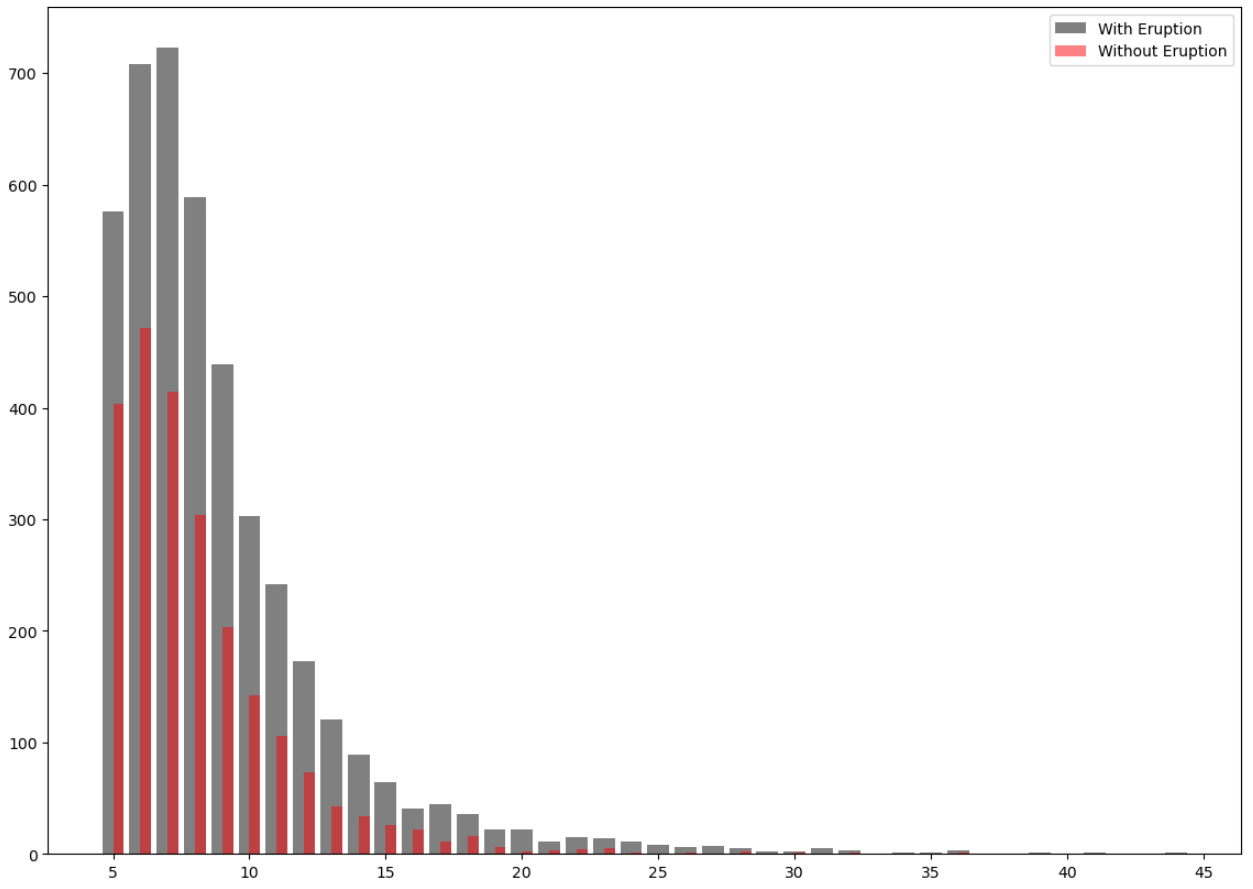}
  \caption{Count of timescales for evolution pathway \#1}
  \label{fig:sub1}
\end{subfigure}
\begin{subfigure}{.8\textwidth}
  \centering
  \includegraphics[width=\linewidth]{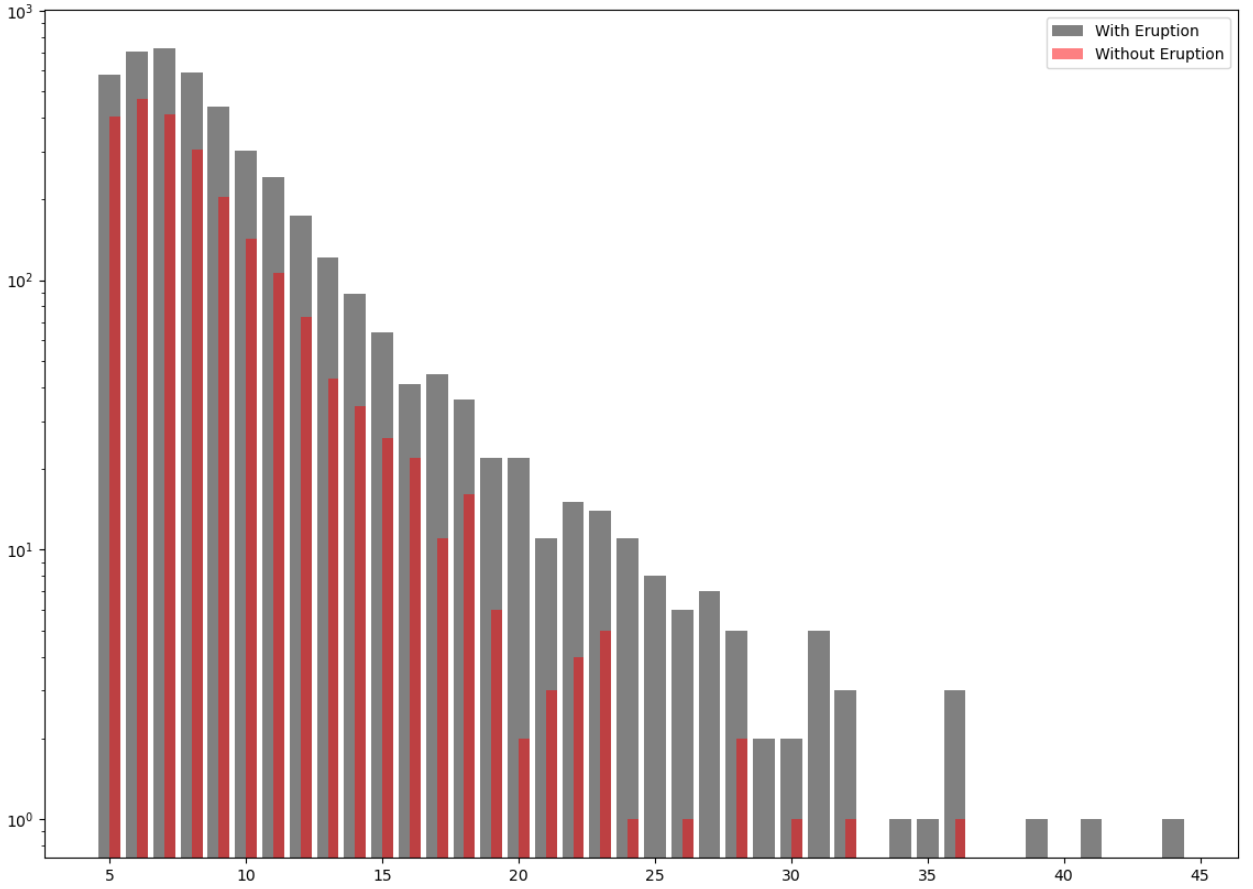}
  \caption{Log plot of the timescales for evolution pathway \#1}
  \label{fig:sub2}
\end{subfigure}
\caption{Plots of the count for cluster evolution pathway \#1 in Table \ref{tab:top_evolutions} with and without eruption, using simulation \#1.}
\label{fig:evolution_timescales}
\end{figure}

Example timescales for cluster evolution pathway \#1 (see Table \ref{tab:top_evolutions}) as it occurs in simulation \#1 are shown in Figure \ref{fig:evolution_timescales}. The top four timescales lie within 5-8 days, though some instances take up to 44 days to evolve. These evolutions represent a localized change in stratospheric temperature due to aerosols. Though these pathway evolutions occur without the eruption, the With Eruption simulations show the gradual increase in the frequency and number of evolution instances, building up over a series of months and staying prevalent for an extended time afterwards. These results confirm the presence of the stratospheric warming pathway in the With Pinatubo data.  
Statistics for the other simulations and the other pathways in Table \ref{tab:top_evolutions} are similar.

It should be noted that the strictness of assertions A1-A3 as well as the maximum evolution length can affect the timescales. For instance, not requiring that the T050 cluster evolution be monotonically non-decreasing would allow for some wavering between T050 clusters, while still ensuring that the final T050 cluster is greater than at the beginning. Additionally, allowing for evolutions greater than length 4 would allow the tracking of evolutions that have more intermediate steps.

 

\begin{figure}[!htb]
 
  \centering
  \includegraphics[scale=0.45]{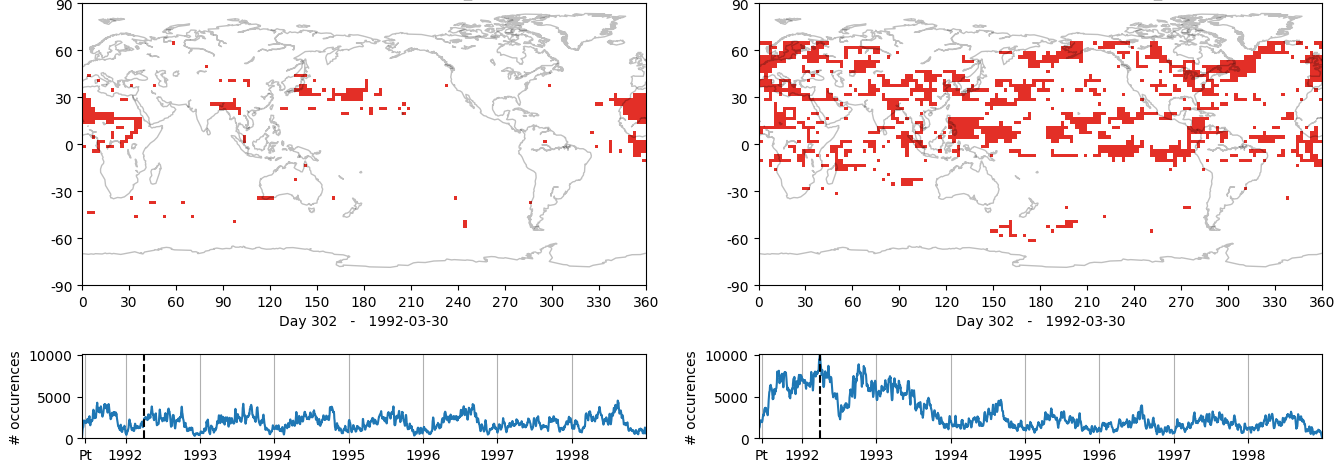}
   \caption{Top panel: visualization of analysis partitions identified as satisfying assertions A1-A3 on March 30, 1992, 302 days after the eruption.  Bottom panel: time-histories of number of pattern occurrences as a function of time.  Left column: Without Pinatubo.  Right column: With Pinatubo. Analysis partitions are marked red in the map when one or more patterns satisfying A1-A3 are occurring at the given instant of time. Vertical dashed line indicates the simulation day of the current snapshot. This result is for a single ensemble member, specifically, ensemble member one.}
  \label{fig:pinatubopatterns}
\end{figure}

Having identified all partition-timesteps that are part of these evolutions, we plot both the spatio-temporal locations and cumulative prevalences 
of the clustering satisfying rules A1--A3 
over time in Figure \ref{fig:pinatubopatterns}.  The top panels in this figure 
show the prevalance over time of partitions satisfying the assertions of interest in the Without Pinatubo (left) and With Pinatubo (right) simulations on March 30, 1992, or day 302 in our simulations. We pick this temporal location, as this is the time at which the pattern prevalences peak (as indicated by the dashed black vertical line in the bottom subpanels of Figure \ref{fig:pinatubopatterns}).  
This result is consistent with the literature, which shows that 
the largest sulfate particles with the highest absorption efficiency of outgoing longwave radiation are measured in 1992, and hence, peak longwave radiative impacts and stratospheric temperature differences are expected in 1992 \cite{Brown:2024}.
The prevalence of the rules A1--A3 seems to follow a seasonal pattern in the Without Pinatubo simulation (left panel of Figure \ref{fig:pinatubopatterns}). In contrast, in the With Pinatubo simulation (right panel of \ref{fig:pinatubopatterns}), there is a huge spike that seems to occur at the end of spring in 1992 which then dips during the summer and returns to its heightened state.  These impacts are attributed to the Mount Pinatubo eruption.  

\begin{figure}[!htb]
 
  \centering
  \includegraphics[scale=0.5]{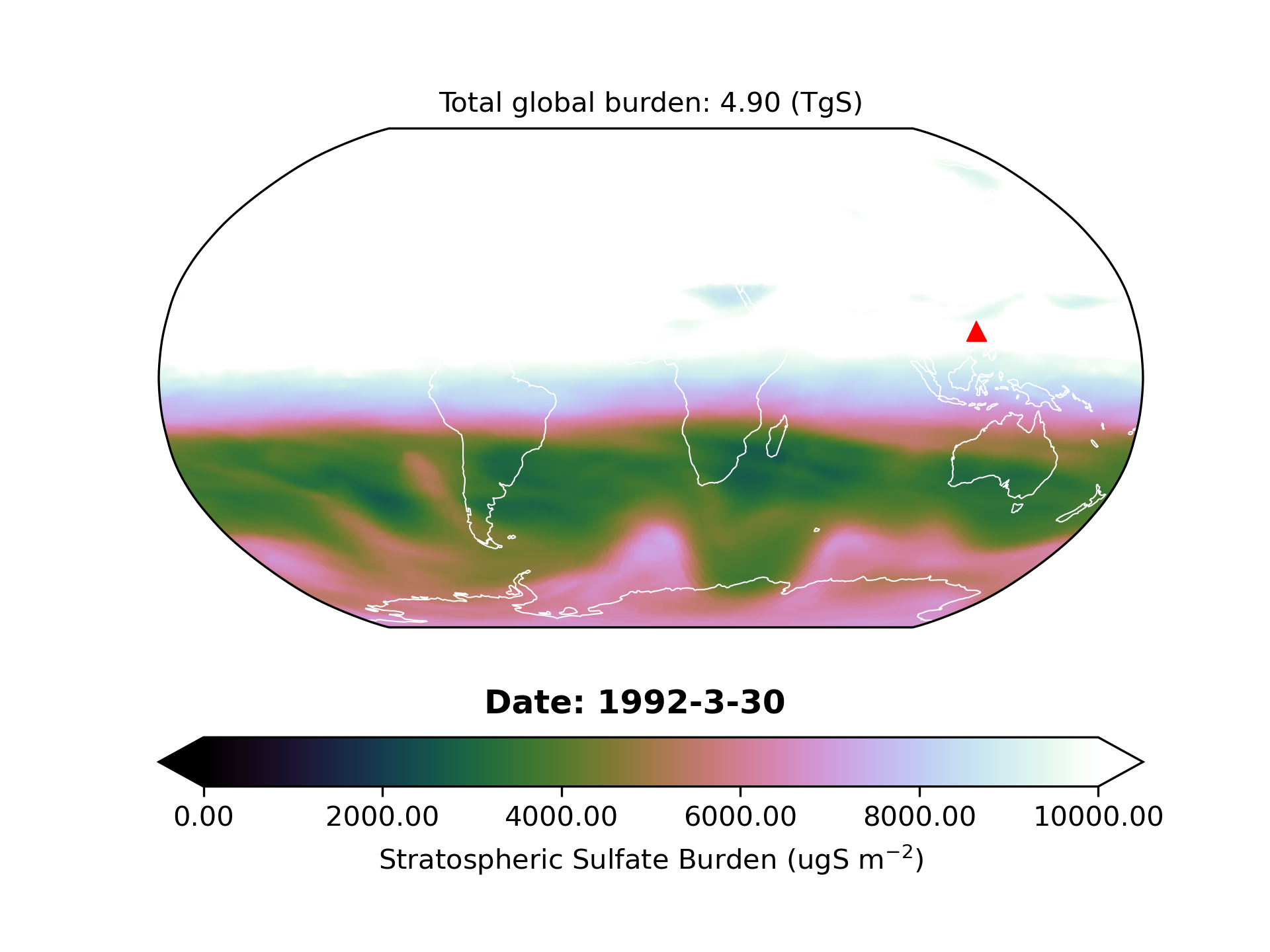}
   \caption{The stratospheric sulfate burden on March 30, 1992.  The location of Mount Pinatubo is marked with a red triangle.}
  \label{fig:strat_burden}
\end{figure}

It is interesting to observe that the locations of the evolution 
patterns identified are mostly in the northern hemisphere.  
Importantly, this result is consistent with both simulations and observations of the Mount Pinatubo eruption. 
 Figure \ref{fig:strat_burden} plots the total stratospheric sulfate burden on March 30, 1992 from our E3SMv2-SPA simulation of the eruption.  
It is clear from this plot that there is far more aerosol present in the northern hemisphere than in the southern hemisphere.  Moreover, since late March marks the start of the transition from winter to spring in the northern hemisphere, there is likely increased solar flux in the northern hemisphere, which emphasizes the relevant effects.

The reader can observe by examining the left panel in Figure \ref{fig:pinatubopatterns} that there are some patterns identified in the Without Pinatubo data.  The patterns seen over Africa are likely due to dust from the Sahara, and the patterns in the southern hemisphere may be attributed to the Cerro Hudson eruption in August 1991 \cite{Case:2017}, as discussed earlier.  

\begin{figure}[!htb]
 
  \centering
  \includegraphics[scale=0.5]{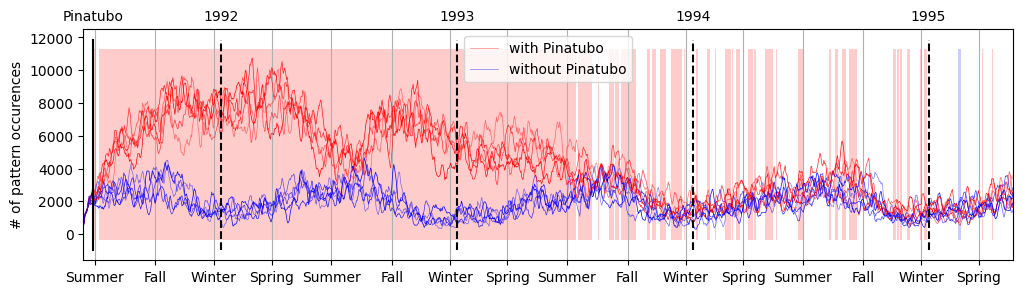}
   \caption{Statistical significance in the first 4 years of the difference in pattern evolution prevalence With Pinatubo vs Without Pinatubo over time. Red curves correspond to the instantaneous number of pattern occurrences for each ensemble member With Pinatubo. Blue curves similarly correspond to Without Pinatubo pattern occurrences. Red shading indicates With Pinatubo occurrences have a statistically-significant increase over Without Pinatubo and blue shading indicates the opposite.}
  \label{fig:ttest_patterns}
\end{figure}

Although the primary impact of the eruption is evident in Figure \ref{fig:pinatubopatterns}, we confirm this qualitative observation by performing a rigorous statistical analysis. Figure \ref{fig:ttest_patterns} shows the results of a $t$-test on the statistical significance of patterns which match assertions A1--A3. This $t$-test is similar to the one shown in Figure \ref{fig:tstat_example}. The null hypothesis is that the prevalence of cluster evolutions that fit A1--A3 are drawn from the same distribution. The red areas are the timesteps where the With Pinatubo simulations contain cluster evolutions that fit A1--A3 significantly higher than what occurs in the Without Pinatubo simulations. Similar to the clustering $t$-test, the With Pinatubo simulations stand out, containing more cluster evolutions that meet assertions the A1--A3 than the Without Pinatubo simulations. However, the signal for the occurrence of the sequences is much stronger and involves multiple variables. This is promising in that the detection of interesting cluster evolution sequences may amplify interesting and relevant signals and allow researchers to detect very subtle effects better than with simple single variable analysis.

We end by emphasizing that, while the plots presented in this section resemble the detection plots shown earlier in Section \ref{sec:clustering_results}, our multivariate mining-based approach is not just identifying when and where the relevant features of interest are statistically different in the With Pinatubo data relative to the Without Pinatubo data; it is actually tracking the prevalence of pathways that meet assertions A1--A3, which define the stratospheric warming pathway of interest herein.  Through the results and discussion above, we have thus demonstrated that our multivariate mining-based approach is capable of confirming hypothesized pathways encoded in the relevant association rules.  


\section{Conclusions and future work} \label{sec:conc}

This paper presents a novel approach for efficiently detecting impacts and tracing spatio-temporal source-impact pathways in climate data using signature-based clustering and NLP-based multivariate data-mining.  This research presents a significant advancement in our ability to find statistically-significant
differences in pattern dynamics within climate simulations.  Our approach has a number of advantages over state-of-the-art detection and impact analysis methodologies: (i) its parameters are selected in a rigorous manner, (ii) it is unsupervised and does not require large amounts of data, (iii) it can be used to extract multivariate relationships, and (iv) it can be used not only to confirm known relationships, but also to discover new ones. 
We verify our approach on an exemplar problem involving the 1991 eruption of Mount Pinatubo in the Philippines, whose impacts on surface and stratospheric temperature are well-characterized.  Specifically, we demonstrate that our approach is capable of detecting and characterizing the evolution of the so-called stratospheric warming pathway.  While the signature-based clustering step of our workflow can be sufficient when looking at variables one at a time, identifying true multivariate relationships and impacts efficiently from the clustered data requires tools from NLP, which are used to mine frequent cluster evolution patterns in a set of locally-defined cluster evolution sequences.  

There are several natural directions worthy of exploration in future work.
Although the multivariable data-mining step of our workflow can be used to discover unknown relationships with no prior domain knowledge, in the results presented in Section \ref{sec:mining_results}, our method is used solely for confirmatory analyses.  There is a great
opportunity to use our NLP-based data-mining capabilities for discovery, 
where one would mine the clustered data for pattern evolution sequences, extract those that are statistically-significant, and present them to climate scientists for interpretation/analysis. 
If the evolution patterns discovered
are unknown to climate scientists but physically plausible, they have the potential to inform climate
science. If the evolution patterns are not physically plausible, our approach can be used to identify
biases or bugs within the climate model used to generate the data used in our analysis. Because of the combinatorial explosion of possibilities, discovery focused analysis could be time consuming from a computational perspective, and might require significant human time and expertise to distill truly new and useful information.

There are additional opportunities to use our approach in either a confirmatory or exploratory setting to study impacts of either the Mount Pinatubo eruption or other climate sources of interest.  For the Mount Pinatubo exemplar, our approach has the potential to provide insight into some possible impacts of the Mount Pinatubo event that are not well-agreed upon, e.g., the northern hemisphere winter warming that occurred in 1991-1992 \cite{Polvani:2019, robockMao1992, parker1996impact, Ehrmann:2024}.  Moreover, we can utilize our method to study more complex source-impact pathways within the context of the Mount Pinatubo exemplar, e.g., the impacts of the eruption on agriculture and crop production \cite{proctor2018, yang2020, rsingh2024, Li:2024}.  Other exemplars involving stratospheric aerosol injection (SAI) that are worth considering in future work whose impacts are of great interest are the 2020 Australian wildfires in \cite{Hirsch:2021}, the 2010 Icelandic volcano eruption \cite{Gudmundsson:2012}, and SAI as a geoengineering strategy in which 1 Tg sulfur dioxide per year may produce on the order of 0.1$^{\circ}$C cooling \cite{Kravitz:2017}. 

In this research, we studied climate simulation data. However there is nothing about our methods that preclude their use on observed/experimental data, as well. In addition, although we have specifically applied our new methodology to the climate domain, the methods, themselves, are not climate specific. We believe that we can apply the methodology presented to many domains. The intended application must only have spatio-temporal dynamics that researchers are wanting to analyze. Therefore, we see a wide range of potential use cases in domains such as fluid dynamics and combustion.


\section*{Data and Code Availability} 
Data from the full E3SMv2-SPA simulation campaign including pre-industrial control, historical, and Mount Pinatubo ensembles will be hosted at Sandia National Laboratories with location and download instructions announced on \url{https://www.sandia.gov/cldera/e3sm-simulations-data/} when available. The core libraries used to create partition signatures for clustering are located at \url{https://github.com/sandialabs/ISML}. This repository will also host the clustering and multivariate pattern mining as a future update.
The E3SMv2-SPA code used to generate the data used in our analyses is also open-source and available at \url{https://github.com/sandialabs/CLDERA-E3SM}.

\section*{Acknowledgements}

This work was supported by the Laboratory Directed Research and Development program at Sandia National Laboratories, a multimission laboratory managed and operated by National Technology and Engineering Solutions of Sandia, LLC, a wholly  owned subsidiary of Honeywell International, Inc., for the U.S. Department of Energy’s National Nuclear Security Administration under contract DE-NA-0003525. 
The research used resources of the National Energy Research Scientific Computing Center (NERSC), a Department of Energy Office of Science User Facility using NERSC award BER-ERCAP0026535.  The writing of this manuscript was funded in part by the third author’s (Irina Tezaur's) Presidential Early Career Award for Scientists and Engineers (PECASE). 

This article has been authored by an employee of National Technology \& Engineering Solutions of Sandia, LLC under Contract No. DE-NA0003525 with the U.S. Department of Energy (DOE). The employee owns all right, title and interest in and to the article and is solely responsible for its contents. The United States Government retains and the publisher, by accepting the article for publication, acknowledges that the United States Government retains a non-exclusive, paid-up, irrevocable, world-wide license to publish or reproduce the published form of this article or allow others to do so, for United States Government purposes. The DOE will provide public access to these results of federally sponsored research in accordance with the DOE Public Access Plan \url{https://www.energy.gov/downloads/doe-public-access-plan}.

The authors would like to thank Hunter Brown, Benjamin Wagman, Ben Hillman, Thomas Ehrmann, and Joe Hollowed for providing invaluable subject matter expertise in stratospheric dynamics and climate modeling, which enabled 
the interpretations discussed within this paper. 

\bibliographystyle{elsarticle-num}
\bibliography{citations}

\end{document}